%% file: acl_latex.tex
\documentclass[11pt]{article}

\usepackage[final]{acl}

\usepackage{times}
\usepackage{latexsym}

\usepackage[T1]{fontenc}
\usepackage[utf8]{inputenc}

\usepackage{microtype}

\usepackage{inconsolata}

\usepackage{graphicx}
\usepackage{algorithm}
\usepackage{subcaption}
\usepackage{algorithmic}
\usepackage{amsmath}
\usepackage{amssymb}
\usepackage{multirow}
\usepackage{booktabs}
\usepackage{adjustbox}
\usepackage{booktabs}
\usepackage[table]{xcolor}

\title{Predictive Bandit Curriculum Reinforcement Learning \\ via Gradient-Based Sample Group Discovery}

\author{
    Liang Xu$^{1,2}$ \quad
    Shufan Shen$^{1,2}$ \quad
    Zhongni Hou$^{3,*}$ \quad
    Yufei Zhang$^{3}$ \quad
    Wei Lin$^{3}$ \quad
    Guojun Yin$^{3}$ \quad
    Shuhui Wang$^{1,*}$ \\[0.5em]
    $^{1}$State Key Lab. of AI Safety, Institute of Computing Technology, \\
    Chinese Academy of Sciences \\
    $^{2}$University of Chinese Academy of Sciences \\
    $^{3}$Meituan \\[0.3em]
}
\author{%
Liang Xu$^\text{1,2}$,\enspace Shufan Shen$^\text{1,2}$,\enspace Zhongni Hou$^\text{3,}$\footnotemark[2],\enspace Chao Bi$^\text{1}$, \\ \textbf{Yufei Zhang}$^\textbf{3}$,\enspace  \textbf{Wei Lin}$^\textbf{3}$,\enspace \textbf{Guojun Yin}$^\textbf{3}$, \textbf{Shuhui Wang}$^\textbf{1,}$\footnotemark[2] \\
\textsuperscript{1}State Key Lab. of AI Safety, Institute of Computing Technology,\\ Chinese Academy of Sciences\\
\textsuperscript{2}University of Chinese Academy of Sciences 
\textsuperscript{3}Meituan \\[2pt]
\texttt{\{xuliang23z\}@ict.ac.cn}
}

\begin{document}
\maketitle
\renewcommand{\thefootnote}{\fnsymbol{footnote}}
\footnotetext[2]{Corresponding authors.}
% \footnotetext[2]{Corresponding author.}
\renewcommand{\thefootnote}{\arabic{footnote}}
\begin{abstract}

%Reinforcement learning (RL) has become a key paradigm for large language model (LLM) post-training, yet its effectiveness strongly depends on how training data are organized and scheduled. Existing curriculum methods commonly construct curriculum units using predefined data properties and estimate their utility from current or recent learning signals. 
Curriculum learning, which organizes training samples according to their learning priority, has been widely adopted in reinforcement learning to improve training efficiency. However, existing methods commonly construct curriculum units using predefined data properties and estimate their utility by simply aggregating a single proxy signal from recent training history.
Such designs may group samples with different optimization effects and fail to anticipate how their utility evolves in relation to the policy. We propose Predictive Gradient-based Bandit (PGB) Curriculum, which first discovers optimization-aware learning units by clustering projected sample-level gradients. PGB then employs a lightweight temporal predictor to forecast the future utility of each unit and converts the predictions into a Boltzmann sampling distribution for adaptive data allocation. Experiments across multiple model scales and reasoning benchmarks show that PGB consistently outperforms standard GRPO, difficulty-based curricula, and history-based scheduling strategies, demonstrating improved data utilization and training efficiency.

\end{abstract}


\section{Introduction}
Reinforcement learning (RL) has emerged as a key paradigm for improving the reasoning abilities of large language models (LLMs)~\cite{openai_reasoning2024,Guo2025DeepseekR1, Schulman2017PPO, shao2024GRPO}.
Despite achieving remarkable performance, 
most RL methods sample prompts from a fixed uniform distribution over the training dataset during training.
Such a plain sampling strategy may lead to inefficient data utilization, \textit{i.e.}, well-learned samples offer diminishing marginal improvements, whereas overly challenging samples may yield indistinguishable rewards and thus weak learning signals~\cite{fang2019curriculum}. Moreover, as the policy evolves, the same sample may transition from being overly difficult to informative and eventually to well learned, indicating that its role in optimization changes throughout training.
% Therefore, % 需要一种
% efficient RL post-training requires identifying and prioritizing samples that provide informative optimization signals for the current policy state. 
% the effectiveness and efficiency of RL post-training depend critically on how training samples are organized and selected throughout optimization. 
%Different training examples contribute unequally to policy improvement, thus simply optimizing over uniformly sampled data can lead to inefficient learning dynamics. The model may overfit to overly simple sampes while failing to sufficiently explore more informative examples, or alternatively struggle with excessively difficult samples that provide weak learning signals. 
% Different samples contribute unequally to policy improvement depending on the current model capability. Uniformly sampling from the entire training set may lead to inefficient learning dynamics: mastered samples offer diminishing marginal improvements, while overly challenging samples may produce sparse or indistinguishable rewards. 
%Therefore, it is essential to select training samples that match the current capability of the policy.

\input{Input/image_intro}

%To improve the training efficiency of RL, recent studies incorporate curriculum learning~\cite{Bengio2009Curriculum} strategies that group training samples and then allocate the sampling probabilities across these groups~\cite{Lin2025CurriculumforRL}.
Recent studies have incorporated curriculum learning~\cite{Bengio2009Curriculum} into RL post-training to dynamically prioritize samples that are most beneficial to the current policy along the learning procedure. A common solution first organizes training samples into predefined groups and then adaptively adjusts the sampling probability of each group throughout optimization~\cite{Lin2025CurriculumforRL}.
%For sample grouping, current methods adopt rule-based strategies to cluster similar samples according to specific metrics
The samples are usually clustered according to predefined attributes or statistics such as difficulty~\cite{campos2021languagemodeling}, success rate~\cite{xi2024training}, and task categories~\cite{teh2017Distral}. 
%For sampling probability allocation, these methods determine the probability of each group by quantifying its contribution to training performance~(e.g., advantage~\cite{wang2025dump}, learning process~\cite{portelas2020teacher}). 
For sampling probability allocation, these methods assign priorities to different groups based on proxy signals such as average advantage~\cite{wang2025dump} or learning progress~\cite{portelas2020teacher}, and adjust their sampling probabilities accordingly.
%By selectively sampling training data, curriculum-based strategies largely enhance RL efficiency compared to the uniform sampling paradigm.
By selectively sampling training data, curriculum-based strategies can substantially improve data utilization and training efficiency over uniform sampling.
% By selectively sampling training data, curriculum-based strategies can improve data utilization over uniform sampling.
%通过有偏好的选择训练数据，基于课程学习的方法相比uniformly sample的范式提升了RL的效率

%However, existing strategies are limited to external metrics while overlooking the intrinsic states of the model. 
%However, existing curriculum learning methods remain limited in both sample grouping and group scheduling.
However, existing curriculum methods primarily assess a sample's learning utility, {\it i.e.}, the contribution of a sample to model learning, using predefined data properties~(e.g., difficulty~\cite{qi2025webrl}) or historical performance signals~(e.g., absolute advantage~\cite{chen2025sec}). They often overlook that the utility is policy-dependent and continuously evolves as policy changes. This issue manifests itself both in sample grouping and in group scheduling.
\textit{First}, existing methods usually group training samples according to predefined properties, such as task categories~\cite{teh2017Distral} or success rate~\cite{xi2024training},
%rather than their contributions to model optimization. 
rather than the optimization effects they induce under a given policy.
However, samples with similar properties do not necessarily induce similar optimization effects, as shown in Figure~\ref{image:intro}(b).
Consequently, groups constructed using such metrics may contain samples with heterogeneous optimization effects, thereby weakening the consistency of each group and reducing the effectiveness of group-level scheduling. 
\textit{Second}, 
the sampling probabilities at the current training step are determined by simply aggregating the historical observations of a single proxy metric~(\textit{e.g.}, average advantages). 
Simple aggregation reduces rich historical observations to a single scalar, making it difficult to capture their evolving patterns and potentially rendering the estimate outdated as the policy evolves. 
Worse still, the reliance on a single proxy metric neglects the rich dynamics of variables such as policy parameters or gradients, and thus provides a deficient characterization of the current policy state. 
%Consequently, such estimation methods may be insufficient to effectively capture how the signal evolves over time, thereby reducing the effectiveness of curriculum scheduling.
Consequently, such estimation methods may be insufficient to capture the temporal evolution of learning signals and their dependence on the policy state, thereby limiting the effectiveness of curriculum scheduling.

In this paper, we propose \textbf{P}redictive \textbf{G}radient-based \textbf{B}andit (PGB) Curriculum for RL-based LLM Post-training, a policy-state-informed curriculum learning framework that 
incorporates policy-dependent optimization information into both sample grouping and curriculum scheduling.
Specifically, for group construction, PGB clusters samples based on their gradient representations rather than external metrics. The gradient representations provide an optimization-aware approximation of how samples influence parameter updates under the given policy, enabling PGB to identify samples with similar learning signals.
For sampling probability allocation, we introduce the predictive multi-armed bandit (MAB). Instead of simply aggregating historical proxy signals to estimate learning utilities, PGB employs a lightweight learnable temporal model to predict these utilities by jointly modeling the evolution of external metrics and intrinsic policy states during optimization. 
Based on the predicted utilities, PGB constructs a Boltzmann sampling distribution that dynamically prioritizes high-utility groups. 
By incorporating the policy's internal states during sample grouping and curriculum scheduling,
PGB better captures the evolving relationship between training samples and the state of the policy, leading to more effective curriculum scheduling.
% enabling more adaptive curriculum scheduling under non-stationary RL dynamics.

Extensive experiments demonstrate that PGB consistently improves the training efficiency and final performance of RL. Compared with the uniform sampling strategy~\cite{shao2024GRPO}, PGB achieves substantial improvements on both in-domain~(ranging from $8.8\%$ to $14.3\%$) and out-of-domain datasets~(ranging from $14.7\%$ to $16.3\%$). 
%The improvement is even more pronounced on the challenging dataset AIME~(ranging from $22.3\%$ to $46.5\%$). 
PGB also substantially improves the training efficiency by achieving competitive performance with fewer steps, reaching the final performance of GRPO earlier during training.
Moreover, compared with existing curriculum learning methods, PGB achieves substantial performance gains~(ranging from $4.9\%$ to $9.2\%$), further demonstrating the effectiveness of our approach. Overall, these results indicate that PGB enables more effective data utilization during training by leveraging the intrinsic states of the model, leading to significant improvements to current RL-based post-training methods.

\section{Related Work}
\subsection{Reinforcement Learning for LLM Post-Training}
%Reinforcement learning~\cite{Guo2025DeepseekR1,lambert2025tulu3} has become a widely adopted paradigm for LLM post-training. PPO~\cite{Schulman2017PPO} and GRPO~\cite{shao2024GRPO} are representative policy optimization algorithms that improve the stability and scalability of RL training.
%Despite their success, the effectiveness of RL-based post-training strongly depends on how training data are organized during training.
%For these RL tasks, training data are highly heterogeneous, where samples vary significantly in difficulty and learning utility~\citep{schaul2016prioritized}. Direct sampling of these data often leads to inefficient learning, as the policy may focus on easy or difficult samples, leading to inefficient and potentially unstable training. To address this issue, curriculum learning has been adopted to improve training efficiency. 
Reinforcement learning~\cite{Guo2025DeepseekR1,lambert2025tulu3}, using algorithms such as PPO~\cite{Schulman2017PPO} and GRPO~\cite{shao2024GRPO}, is widely used for LLM post-training.
Its effectiveness depends on data allocation, as samples vary in difficulty and learning utility~\citep{schaul2016prioritized}.
Curriculum learning improves training efficiency by adaptively prioritizing useful samples.

\subsection{Curriculum Learning}
Curriculum learning~\cite{Bengio2009Curriculum} aims to improve training efficiency and final performance by controlling the order or distribution of training data during model optimization. 
In reinforcement learning, curriculum design generally involves two complementary components: sample grouping and curriculum scheduling. 

\input{Input/image_overview.tex}

According to the granularity at which curriculum decisions are made, existing methods can be broadly categorized into task-based and sample-based approaches. 
Task-based curriculum reinforcement learning (CRL)~\cite{huang2022CAMRL} treats tasks or task configurations as scheduling units and controls their learning order~\cite{Lin2025CurriculumforRL}. Representative methods include Reverse Curriculum Generation~\cite{Florensa17Reverse}, TSCL~\cite{matiisen2019teacher}, and ALP-GMM~\cite{portelas2020teacher}, which select or generate tasks according to difficulty or learning progress. Recent works have extended this idea to LLM agents and reasoning tasks, such as R3~\cite{xi2024training} and WebRL~\cite{qi2025webrl}. 
Compared with task-level scheduling, sample-based approaches allow more flexible control over the training distribution by composing batches from different groups.
Representative sample-based approaches include Prioritized Experience Replay  (PER)~\cite{schaul2016prioritized}, Experience Replay Optimization (ERO)~\cite{zha2019experience}, and CHER~\cite{fang2019curriculum}, which prioritize training samples according to estimated learning value or experience utility.
Several sample-based methods formulate curriculum scheduling as a multi-armed bandit problem, where predefined data groups are treated as arms and selected according to estimated learning utility. SEC~\cite{chen2025sec} models curriculum selection under a non-stationary bandit framework using absolute advantage as the learning signal, while DUMP~\cite{wang2025dump} employs a UCB-based scheduler with recent statistics to allocate sampling budgets across data groups.
Gradient information also supports training-data influence estimation and subset selection. Influence functions~\cite{koh2017understanding} and TracIn~\cite{pruthi2020estimating} estimate how training examples affect model predictions, while CRAIG~\cite{mirzasoleiman2020coresets} and GRAD-MATCH~\cite{killamsetty2021grad} select weighted subsets that approximate target gradients. These methods characterize sample influence or gradient representativeness, but do not explicitly predict the future learning utility of curriculum groups during RL post-training.

Existing bandit-based curricula typically operate on predefined categories or data distributions, whose samples may induce heterogeneous optimization effects. Although their utility estimates adapt to training feedback, they generally rely on current or recent proxy statistics without explicitly modeling temporal patterns and evolving policy states. Consequently, these estimates may lag behind changes in learning utility during non-stationary RL training.
%Despite their effectiveness, existing MAB-based curriculum methods for LLM post-training generally construct arms using predefined data properties or observed performance statistics, such as task type or difficulty, although such criteria may not reflect whether samples induce similar learning signals or optimization effects. Moreover, they typically estimate the future utilities using current or recent observations, without explicitly modeling how the utility evolves with the changing policy state. Since RL training is inherently non-stationary, such retrospective estimates may fail to anticipate future learning benefits. These limitations motivate more advanced curriculum methods that can uncover the latent optimization structure of training data and explicitly predict the future utility of curricula.

%here
\section{Method}

In this section, we present the PGB framework, which incorporates model-internal optimization information into both data organization and curriculum scheduling during training. The framework is designed to address two key limitations of existing methods, {\it i.e.}, the reliance on external properties for data grouping and the use of retrospective signals for curriculum scheduling. The overall framework consists of two components: (i) gradient-based learning unit discovery and (ii) predictive multi-armed bandit CRL.

\subsection{Preliminaries}

We begin by motivating the gradient-based representation used for learning unit discovery. Inspired by the Neural Tangent Kernel (NTK) theory~\cite{jacot2018neural}, we first establish the connection between sample-wise gradients and local optimization effects, which provides the foundation for grouping the training data in the first stage. For a training sample $(x_i,y_i)$, we represent its local update signal using the gradient $g_i$ of its training loss:
\begin{equation}
    g_i=\nabla_\theta \ell_i(\theta).
\end{equation}
Consider a one-step gradient update induced by sample $i$:
\begin{equation}
    \theta'=\theta-\eta g_i.
\end{equation}
By first-order Taylor expansion, the loss change of another sample $j$ is approximated as
\begin{equation}
    \ell_j(\theta')-\ell_j(\theta)
\approx -\eta g_j^\top g_i.
\end{equation}
Therefore, $g_j^\top g_i$ characterizes how an update induced by sample $i$ locally affects the loss of sample $j$. Since the raw inner product combines gradient magnitude and direction, and the magnitude may reflect sample-specific scale, we normalize the gradients and focus on their directional alignment when constructing learning units.

\subsection{Gradient-based Learning Unit Discovery}

Based on the local gradient analysis above, we group training samples with similar gradient directions into learning units~\cite{Caeiro2014PoEML}. We use supervised fine-tuning (SFT) gradients computed on reference trajectories under the initial policy as a proxy for sample-level update directions, enabling gradient-based grouping without additional on-policy rollouts.
%Since obtaining reliable sample-level RL gradients before training requires costly and stochastic rollouts, we instead use supervised fine-tuning (SFT) gradients as a stable and rollout-free surrogate. Although the two gradients differ in their weighting, both are built upon the trajectory log-likelihood gradient, making SFT gradients an optimization-aware representation for learning unit construction.

Concretely, for each training sample $(x_i,y_i)$, we compute its gradient representation using SFT loss:
\begin{equation}
    \mathcal{L}_{\text{SFT}}(x_i, y_i) = -\frac{1}{T} \sum_{t=1}^{T} \log p_\theta(y_t \mid x_i, y_{<t}),
\end{equation}
where $p_\theta$ denotes the conditional distribution defined by the policy model, $T$ is the number of response tokens in the reference trajectory, $x_i$ is the input prompt, and $y_i$ denotes the corresponding reference trajectory provided by the dataset or generated by a teacher model. The gradient representation of each sample is then defined as:
\begin{equation}
    g_i = \nabla_\theta \mathcal{L}_{\text{SFT}}(x_i, y_i),
\end{equation}
which serves as an optimization-aware representation of the local update signal.
%We then perform clustering to identify learning units:
Since $g_i$ contains both magnitude and directional information, we apply $L_2$ normalization to isolate the direction of the induced model update:
\begin{equation}
    \bar{g}_i=\frac{g_i}{\left\| g_i \right\|_2}
\end{equation}
In principle, learning units can be constructed by directly clustering the normalized gradient representations according to their directional similarity. However, operating in the full gradient space is computationally expensive for large language models.
We therefore apply a Gaussian random projection (GRP), which approximately preserves pairwise distances and similarities in a lower-dimensional space according to the Johnson--Lindenstrauss lemma~\cite{johnson1984extensions}.
Specifically, we sample
\begin{equation}
\begin{aligned}
    \mathbf{R}&\in\mathbb{R}^{d\times D},\\
\mathbf{R}_{ab}&\sim\mathcal{N}\!\left(0,\frac{1}{d}\right),
\end{aligned}
\end{equation}
and compute the projected representation as $z_i=\mathbf{R}\bar{g}_i$.
% \begin{equation}
%     z_i=\mathbf{R}\bar{g}_i
% \end{equation}
According to the properties of GRP, we have
\begin{equation}
    \mathbb{E}\!\left[\langle z_i,z_j\rangle\right]
=
\mathbb{E}\!\left[\langle \mathbf{R}\bar{g}_i,\mathbf{R}\bar{g}_j\rangle\right]
=
\langle \bar{g}_i,\bar{g}_j\rangle,
\end{equation}
which indicates that the projected inner product is an unbiased estimator of the cosine similarity between the original gradients. A detailed theoretical justification is provided in the Appendix.
Therefore, rather than clustering in the original gradient space, we apply spherical $K$-means in the projected space to construct the learning units:
\begin{equation}
\begin{gathered}
    c_i = \text{Assign}(z_i)\in \{1,\ldots,K\},\\
    u_k = \{(x_i,y_i)\in\mathcal{D}\mid c_i=k\}.
\end{gathered}
\end{equation}
We compute the sample-wise gradients using the initial policy before RL training and construct the learning units only once. The fixed units provide a stable partition based on the initial optimization relationships among samples and preserve consistent arm identities throughout training. Their stage-dependent learning utilities are then dynamically modeled by the predictive scheduler.

\subsection{Predictive Bandit Curriculum Learning}
After constructing the learning units, 
the remaining challenge is to allocate the training budget according to their evolving learning utility during optimization.
Since the utility of each unit is policy-dependent and changes as the policy evolves, directly relying on current or historical observations may provide inaccurate estimates of future utility. Therefore, curriculum scheduling can be formulated as a sequential decision-making problem. We model each learning unit as an arm in a multi-armed bandit framework and predict its future utility to guide adaptive sampling.
%Curriculum scheduling is therefore naturally formulated as a sequential decision-making problem. We model this process as a multi-armed bandit problem, where each learning unit corresponds to an arm.

Specifically, given the set of learning units $\mathcal{U}$ obtained by clustering,
\begin{equation}
    \mathcal{U}=\left\{  u_1,u_2,\ldots,u_K \right\}
\end{equation}
we formulate each unit $u_k \in \mathcal{U}$ as an arm in the bandit. Unlike the classical single-pull bandit settings, which select a single arm at each decision step, we consider a probabilistic multi-arm allocation setting where each training iteration distributes a fixed sample budget across multiple arms.
During training, each arm receives a reward signal that quantifies the contribution of its corresponding learning unit.
Following prior work~\cite{xiong2025hsstar}, we refer to this contribution as the learning utility of the unit. Drawing on existing curriculum learning methods~\cite{chen2025sec,wang2025dump}, we use observable reward signals to characterize this utility from two perspectives:
(1) the average absolute advantage, serving as a proxy for the strength of the learning signal
\begin{equation}
    \mathcal{R}_k^{t}
=
\frac{1}{|B_k^{t}|}
\sum_{x_i\in B_k^{t}}
\left|A_i^{t}\right|,
\end{equation}
and (2) reward progress, which captures the temporal changes in model performance
\begin{equation}
    \mathcal{P}_k^t =r_k^t-r_k^{t-1}
\end{equation}
Here, $B_k^{t}$ is the batch sampled from $u_k$, $\left|A_i^{t}\right|$ is the absolute advantage of $x_i$, and $r_k^t$ denotes the average rollout reward. 

%In traditional multi-armed bandit-based curriculum learning methods\cite{chen2025sec,wang2025dump}, the selection of learning units is typically performed based on the historical reward of each arm. However, such methods may not accurately reflect the current learning utility of each learning unit as the reinforcement learning environment is non-stationary. What's more, directly using historical observations as predictions of future rewards may make the sampling strategy highly sensitive to noisy and non-stationary reward signals during training, as shown in Figure \ref{image:unit_prob}.
Existing bandit-based curriculum methods \cite{chen2025sec,wang2025dump} estimate unit utility by aggregating historical proxy signals, without explicitly modeling its evolution with the policy state.
Such estimates may lag behind changing learning dynamics and yield unstable sampling decisions (Figure~\ref{image:predict}).
To address this issue, we explicitly model and predict the future utility of each learning unit, enabling more informed curriculum scheduling decisions. Specifically, we represent the learning state of each unit $u_k$ at step $t$ as
\begin{equation}
    \mathbf{s}_k^t= \left[
    \mathcal{R}_k^t,\;
    \mathcal{P}_k^t,\;
    \mathcal{N}_k^t,\;
    \mathbf{G}_k^t
    \right]
\end{equation}
where $\mathcal{R}_k^t$ is the average absolute advantage, $\mathcal{P}_k^t$ is the reward progress, $\mathcal{N}_k^t$ is the number of samples drawn at the current step, and $\mathbf{G}_k^t$ is the average gradient feature during RL optimization. Unlike the gradients used for unit construction, $G_k^t$ records compact last-layer gradient statistics to characterize the evolving optimization state.
For units that are not sampled, we set $\mathcal{N}_k^t = 0$, and retain $\mathcal{P}_k^t$ and $\mathcal{R}_k^t$ from the previous step.
Instead of estimating future utilities solely from historical rewards, we use the historical learning state trajectories of each unit to predict its future utility. For each learning unit $u_k$, we maintain a sequence of learning states
\begin{equation}
    \mathcal{H}_k^t=
\{\mathbf{s}_k^{t-L+1},\ldots,\mathbf{s}_k^t\},
\end{equation}
and the future utility $\hat{R}_k^{t+1}$ is predicted by 
\begin{equation}
     \left[
    \hat{R}_1^{t+1},
    \ldots,
    \hat{R}_K^{t+1}
    \right]^\top=f_{\phi}\left(
    \mathcal{H}_1^t,
    \ldots,
    \mathcal{H}_K^t
    \right),
\end{equation}
Here, we parameterize $f_\phi$ as a lightweight learnable predictor consisting of a Gated Recurrent Unit (GRU) followed by a Multi-Layer Perceptron (MLP). It jointly processes the historical states of all $K$ learning units to capture temporal dynamics and cross-unit correlations, and predicts their utilities at the next training step. The predictor operates on compact learning-state representations rather than full model parameters or sample-level gradients.
%With the predicted utility $\hat{R}_k^{t+1}$, we compute the curriculum sampling distribution using a Boltzmann policy:
%Unlike the classical single-pull MAB setting, each training iteration allocates a fixed sample budget across multiple arms. 
The predicted utilities $\hat{R}_k^{t+1}$ are converted into allocation probabilities through a Boltzmann policy.
\begin{equation}
\label{equation:Boltzmann}
    n_k^{t+1}=b\pi_k^{t+1}
    =
    \frac{
    b\cdot \exp\left(\hat{R}_k^{t+1}/\tau\right)
    }{
    \sum_{j=1}^{K}\exp\left(\hat{R}_j^{t+1}/\tau\right)
    }
\end{equation}
where $\tau$ is the temperature hyperparameter controlling the exploration--exploitation trade-off and $b$ is the batch size. 

\input{Input/table_main}

Overall, our framework follows a two-step training procedure. \textbf{Step 1: Warm up.} The warm-up step serves two purposes: ensuring sufficient exploration of all learning units and collecting initial state–utility trajectories for predictor training. During this step, training batches are formed by uniformly sampling from the learning units. After each reinforcement learning update, we record the learning state and observed utility of each sampled unit, thereby constructing the initial training sequences for the predictor. 
The predictor is initialized through supervised training on the state–utility trajectories by minimizing the mean squared error between predicted and observed utilities:
%The predictor is then trained in a supervised manner by minimizing the mean squared error between its predicted utilities and the corresponding observed utilities:
\begin{equation}
\label{equation:pred_loss}
    \mathcal{L}_{\mathrm{pred}}
=
\frac{1}{K}
\sum_{k=1}^{K}
\left(
\hat{R}_k^{t+1}
-
R_k^{t+1}
\right)^2.
\end{equation}
\textbf{Step 2: Predictive curriculum RL.} After the warm-up step, the trained predictor is used to predict the future learning utility of each unit at the next training step. Based on the predicted utilities, PGB computes the Boltzmann sampling distribution in \eqref{equation:Boltzmann} and samples data from the units to construct the training batches for reinforcement learning. During training, newly collected learning states are continuously added to the history trajectories, and the predictor is periodically updated according to \eqref{equation:pred_loss} every $T_{pred}$ training iterations to adapt to the evolving learning dynamics. 
The training procedure is shown in Algorithm \ref{alg:pmab}.

\section{Experiments}
This section evaluates the effectiveness of PGB for RL-based LLM post-training across multiple model scales and both in-distribution and out-of-distribution benchmarks. We further conduct ablation studies to analyze the contribution of each component, including gradient-based learning unit construction and predictive utility modeling.

\subsection{Experimental Setup}
We conduct RL-based post-training experiments on the DAPO-Math-17K dataset~\cite{yu2025dapo}. We randomly hold out $1{,}000$ samples for in-domain evaluation and use the remaining samples for training.
We evaluate our method on three Qwen3 model scales~\cite{yang2025qwen3technicalreport}, including Qwen3-1.7B, Qwen3-4B-Thinking, and Qwen3-8B. The evaluation benchmarks include DAPO-Math for in-distribution evaluation and AIME 2024--2026, MATH-500~\cite{hendrycks2021math}, and AMC for out-of-distribution evaluation~\cite{shao2024GRPO,lewkowycz2022solving}, covering a range of mathematical reasoning difficulties.

%For reinforcement learning fine-tuning on reasoning tasks, we conduct the widely-used GRPO~\cite{shao2024GRPO} algorithm as the base training algorithm. We compare PGB against standard GRPO method and a curriculum method based on difficulty ordering. All methods are trained under the same RL settings for a fair comparison. Detailed experimental settings are provided in the Implementation Details section of the Appendix.
We use Group Relative Policy Optimization (GRPO) \cite{shao2024GRPO} as the base RL algorithm and compare PGB with the following baselines: (1) \textbf{GRPO}, which uniformly samples training prompts; and (2) \textbf{Difficulty-UCB}, which reproduces the DUMP curriculum strategy~\cite{wang2025dump}. Specifically, it partitions training samples into groups according to the initial policy's response accuracy and dynamically allocates the sampling budget among these groups using a UCB-based scheduler. We reimplement this method within our training framework. All methods share the same GRPO settings, training budget, and evaluation protocol for fair comparison. Detailed settings are provided in the Appendix.

\subsection{Main Results}
Table \ref{tab:results} shows the main results on both in-distribution (ID) and out-of-distribution (OOD) benchmarks across Qwen3-1.7B, 4B, and 8B. In general, PGB consistently outperforms standard GRPO and difficulty-UCB curriculum baselines on the ID benchmark and in mean OOD performance across all three model variants, supporting the overall effectiveness of our proposed predictive curriculum learning framework.

In ID evaluation (DAPO-Math), PGB achieves consistent improvements over GRPO across all model sizes, achieving relative performance gains of $8.8\%$ to $14.3\%$. These results show that PGB improves the performance of in-domain reasoning across model scales, consistent with its objective of adapting data allocation to the evolving policy state.
PGB also achieves substantial gains on OOD benchmarks, with relative improvements of $14.7\%$--$16.3\%$. The gains are particularly pronounced on AIME 2024--2026, where PGB substantially outperforms standard GRPO and consistently improves the mean AIME score across all model scales, indicating improved average performance on challenging mathematical reasoning problems. PGB also achieves the best or comparable performance on Math-500 and AMC, suggesting that its benefits extend across multiple mathematical reasoning benchmarks. Notably, it consistently surpasses difficulty-UCB curriculum methods by $4.9\%$--$9.2\%$.
Since both methods dynamically allocate the training budget across data groups, the results show that the overall PGB design performs better than the combination of difficulty-based grouping and UCB-based scheduling. The contributions of learning unit construction and predictive scheduling are examined separately in the following ablations.

%Overall, these results demonstrate that PGB enables more effective data utilization in RL-based post-training by grouping the data according to the optimization characteristics and dynamically adapting curriculum decisions based on predicted learning utility, leading to consistent improvements across settings.
Overall, these results demonstrate the effectiveness of combining optimization-aware learning unit construction with predictive curriculum scheduling, leading to consistent improvements across model scales and benchmarks.

\input{Input/image_training_curve}
\input{Input/table_training_curve}
\subsection{Ablation Studies}
\subsubsection{Training Efficiency Comparison} 
To evaluate the training-step efficiency of PGB relative to standard GRPO, we evaluate PGB and standard GRPO by tracking the mean training reward over optimization steps and the accuracy of intermediate checkpoints on the DAPO test set, as shown in Figure \ref{image:training_curve} and Table \ref{tab:training_cruve}. First, as shown in Figure \ref{image:training_curve}, PGB maintains generally higher training rewards than GRPO over most of the training process for both Qwen3-4B and Qwen3-8B models. More importantly, PGB reaches the peak performance of the GRPO baseline at a much earlier training stage (approximately within 100--200 steps), suggesting faster empirical convergence.
Second, from the step-wise evaluation in Table \ref{tab:training_cruve}, PGB consistently outperforms GRPO at every training stage across both model scales. At only 100 training steps, PGB already matches or exceeds the accuracy achieved by standard GRPO at the end of training, further demonstrating improved sample efficiency. 
Finally, the performance gap between PGB and GRPO is maintained throughout training, suggesting that PGB not only accelerates convergence but also achieves higher performance under the same steps.
Overall, these results show that PGB substantially improves training efficiency by enabling faster convergence and more effective utilization of training steps.

\input{Input/table_unit_construct}
\input{Input/table_sampling}

\subsubsection{Effect of Learning Unit Construction} To investigate the impact of different learning unit construction methods on model performance, we conduct comparative experiments with the following methods varying only the method used to construct learning units: (1) \textbf{GRPO}, which performs standard RL training with uniform sampling without constructing learning units;  (2) \textbf{Difficulty}, which partitions training samples into learning units according to the model's response accuracy; (3) \textbf{Rubric}, which constructs learning units based on predefined reasoning skill rubrics; and (4) \textbf{Gradient}, our proposed method. 
As shown in Table \ref{tab:construct}, different learning unit construction strategies have substantial impacts on model performance. Both difficulty- and rubric-based construction generally outperform standard GRPO, suggesting that group-based curriculum scheduling can improve over uniform sampling. Gradient-based grouping achieves the best overall performance, including gains in both DAPO-Math and AIME, indicating that optimization-aware grouping provides a better basis for curriculum construction.
These results demonstrate that gradient features provide a more informative signal for uncovering the learning structure of training data than difficulty levels or reasoning rubrics. Consequently, gradient features provide a more effective basis for constructing learning units for curriculum scheduling.

\input{Input/image_unit_prob}

\subsubsection{Effect of Predictive Scheduling} 
%We compare PGB with two representative historical-reward-based curriculum scheduling strategies, namely Boltzmann sampling and UCB sampling, to evaluate the effectiveness of the predictive scheduling module. As shown in Table~\ref{tab:sampling}, predictive scheduling consistently achieves the best performance across all reasoning benchmarks. Specifically, PGB improves DAPO-Math accuracy from $57.0\%$ and $57.2\%$ under Boltzmann and UCB scheduling, respectively, to $60.8\%$. More pronounced gains are observed on other datasets, such as AIME and AMC, demonstrating the effectiveness of predictive scheduling across benchmarks with varying reasoning difficulty.
We compare predictive scheduling with an SEC-style scheduler~\cite{chen2025sec}, which applies Boltzmann sampling to EMA estimates of absolute advantage, and a UCB scheduler. All strategies use the same gradient-based learning units to isolate the effect of scheduling. As shown in Table~\ref{tab:sampling}, predictive scheduling consistently achieves the best performance in all the benchmarks evaluated, supporting the effectiveness of incorporating future utility estimation into curriculum scheduling.
To better understand this improvement, Figure~\ref{image:predict}(a) visualizes the sampling probability trajectories under predictive and UCB-based scheduling. UCB exhibits frequent fluctuations in allocation probabilities, suggesting its sensitivity to changing reward signals, whereas predictive scheduling produces smoother trajectories. Figure~\ref{image:predict}(b) compares different utility prediction strategies. The GRU-based predictor achieves lower prediction errors than all baselines compared after the initial stage, suggesting that modeling the evolving utility of learning units provides more accurate utility estimates.

%To further understand how predictive scheduling affects curriculum behavior, we visualize the sampling probability trajectories of several representative learning units under predictive and UCB scheduling in Figure \ref{image:unit_prob}. We observe that UCB scheduling exhibits frequent fluctuations with large variations in sampling probability, reflecting its sensitivity to noisy and non-stationary reward signals during RL training. In contrast, predictive scheduling is less sensitive to short-term reward fluctuations by modeling the temporal evolution of learning utility. This behavior suggests that PGB can better capture the underlying learning dynamics and avoid over-reacting to short term reward fluctuations. These results indicate that explicitly predicting future learning utility provides a more informative curriculum signal than directly relying on current or historical reward statistics.

\subsubsection{Generalization to Science Reasoning}
While our main experiments focus on mathematical reasoning tasks, we further evaluate the generalization ability of PGB on SciKnowEval~\cite{feng2025sciknoweval}, covering Chemistry, Biology and Physics questions. We compare PGB with the original model and standard GRPO to investigate whether predictive curriculum scheduling remains effective beyond math reasoning scenarios. As shown in Table \ref{tab:science}, PGB consistently improves the performance across different science reasoning domains. These results provide evidence that PGB can extend beyond mathematical reasoning.

\section{Conclusion}

We present PGB, a model-aware predictive curriculum framework for RL-based LLM post-training. 
%PGB first constructs optimization-aware learning units by clustering projected sample-wise gradients, and then jointly predicts the future utilities of all learning units to dynamically allocate the training budget. 
PGB combines an optimization-aware partition constructed under the initial policy with predictive modeling of how unit utility evolves during RL training.
Experiments across multiple variants of the Qwen3 model and mathematical reasoning benchmarks show that PGB achieves stronger overall ID and OOD performance over standard GRPO and Difficulty-UCB. Further analyses support gradient-based grouping and show that utility prediction improves estimation accuracy and smooths sampling allocation. These results highlight the benefits of integrating model-internal optimization signals for data organization with predictive modeling of evolving learning utility for RL post-training. 

\section*{Limitations}
Our evaluation is limited to the Qwen3 model family and focuses primarily on mathematical reasoning. To partially mitigate this limitation, we evaluate three Qwen3 model scales and include science reasoning benchmarks; generalization to other model families remains untested. Due to the computational cost of RL post-training, we conduct one training run per configuration. To make these single-run comparisons more controlled, we use the same GRPO settings, training budget, and evaluation protocol across methods; variability across runs remains unmeasured.


\bibliography{custom}
\newpage
\appendix

\input{Input/appendix}

\end{document}

%% file: Input/image_intro.tex
\begin{figure}[t]
    \centering
    \vspace{-0.2cm}
    \begin{subfigure}{\linewidth}
        \centering
        \includegraphics[width=0.96\linewidth]{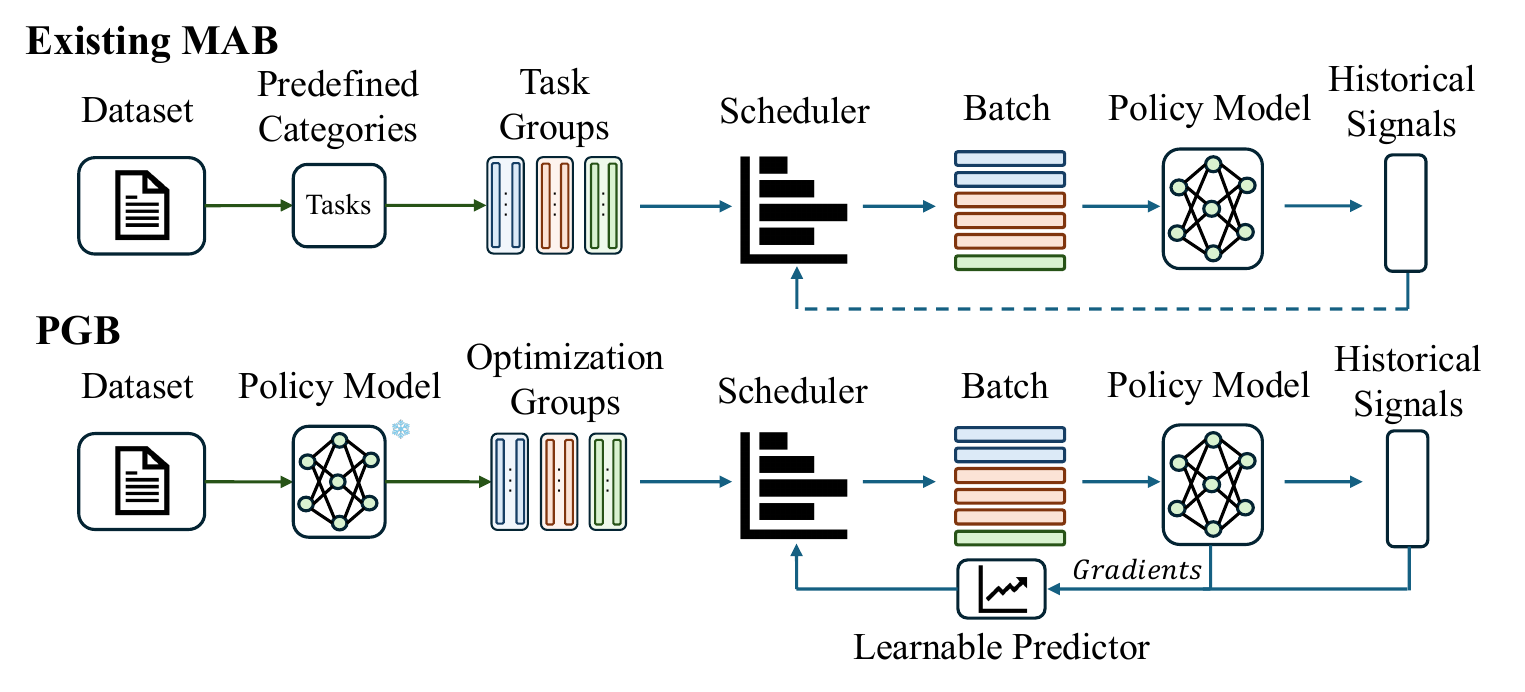}
        \vspace{-0.3cm}
        \caption{Comparison between existing MAB-based CRL and PGB.}
        \label{fig:comparison}
    \end{subfigure}

    \begin{subfigure}{\linewidth}
        \centering
        \includegraphics[width=0.88\linewidth]{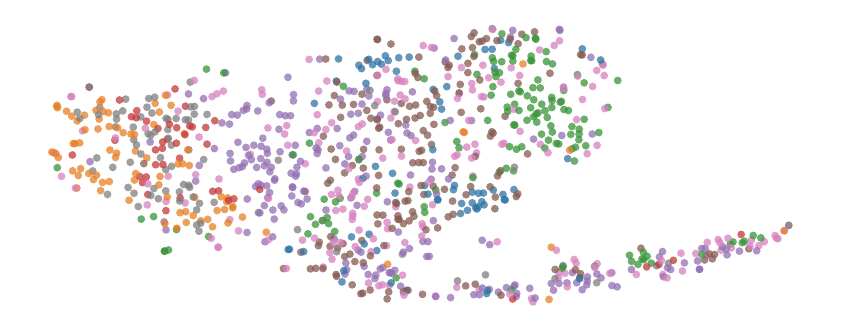}
        \vspace{-0.2cm}
        \caption{Gradient-space distribution of predefined task categories.}
        \label{fig:gradient}
    \end{subfigure}
    \vspace{-0.5cm}
    \caption{(a) Comparison between existing curriculum methods and PGB. 
(b) Task groups in gradient space. Each point is a sample, colored by task category.}
    
    \label{image:intro}
    \vspace{-0.4cm}

\end{figure}

%% file: Input/image_overview.tex
\begin{figure*}[htbp]
%\vskip -0.33in
\begin{center}
\centerline{\includegraphics[width=0.9\textwidth]{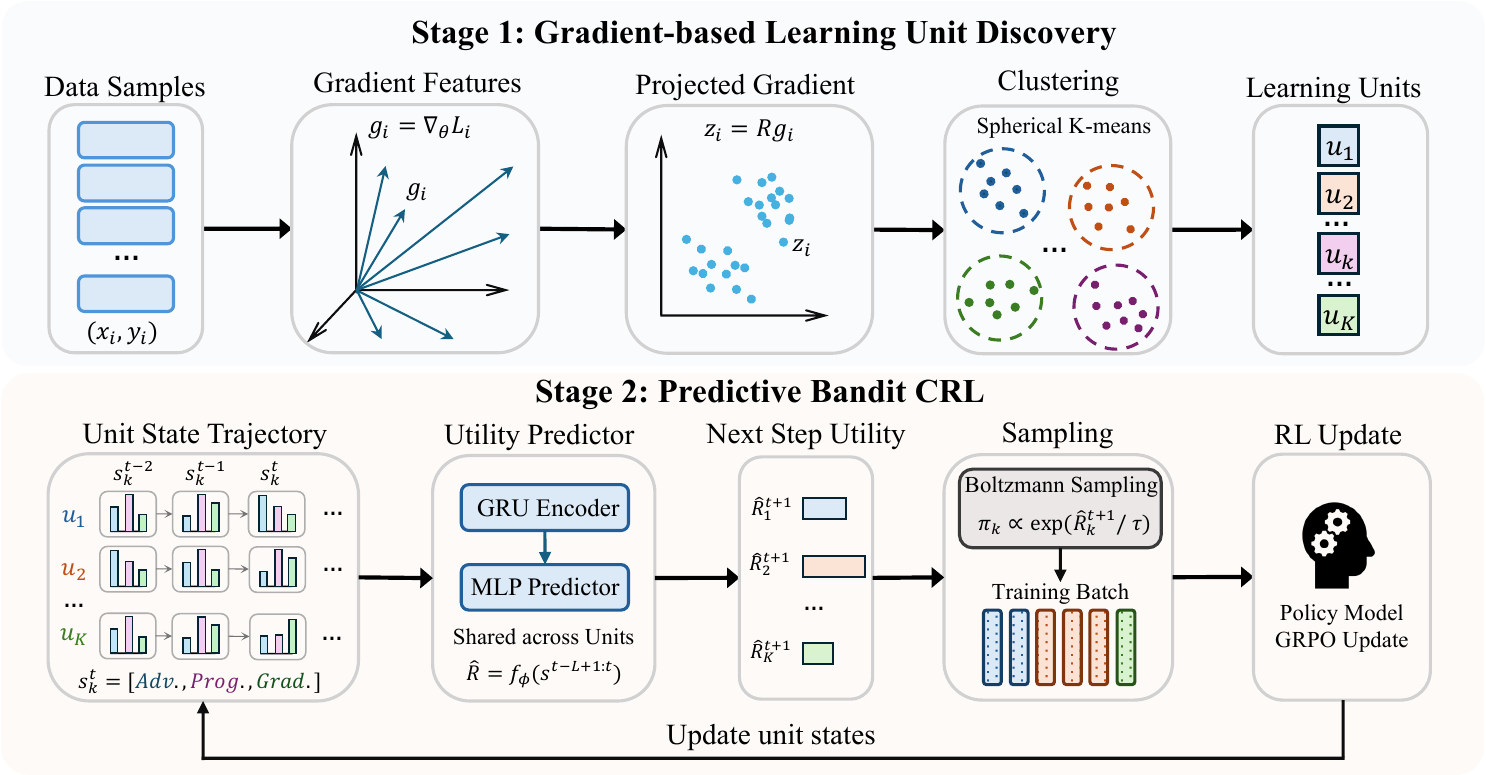}}
\vskip -0.1in
\caption{Overview of our proposed Predictive Gradient-based Bandit (PGB) Curriculum method. PGB first discovers gradient-based learning units according to optimization effects and then performs curriculum scheduling by predicting learning utility from evolving policy states within a multi-armed bandit framework.}
\label{image:overview}
\end{center}
\vskip -0.3in
\end{figure*}

%% file: Input/table_main.tex
\begin{table*}[t]
\centering
\normalsize
\setlength{\tabcolsep}{4pt}
\resizebox{\textwidth}{!}{%
\begin{tabular}{l l c c c c c c c c}
\toprule

\multirow{2}{*}{Model} & \multirow{2}{*}{Method} & ID & \multicolumn{7}{c}{OOD} \\
 \cmidrule(lr){3-3} \cmidrule(lr){4-10}

 & & DAPO-Math & AIME 24 & AIME 25 & AIME 26 & Mean AIME & MATH-500 & AMC & Mean \\
\midrule

\multirow{4}{*}{Qwen3-1.7B}
& Original   & 0.121     & 0.000 & 0.067 & 0.033 & 0.033 & 0.282 & 0.150 & 0.104 \\
& GRPO       & 0.559 & 0.233 & 0.233 & 0.200 & 0.222 & \textbf{0.770} & 0.700 & 0.426 \\
& Difficulty-UCB & 0.565 & 0.267 & 0.233 & 0.233 & 0.242 & 0.742 & \textbf{0.800} & 0.454 \\
\rowcolor{gray!20}
& PGB (Ours)   & \textbf{0.608} & \textbf{0.367} & \textbf{0.300} & \textbf{0.300} & \textbf{0.322} & 0.710 & \textbf{0.800} & \textbf{0.495} \\

\midrule

\multirow{4}{*}{Qwen3-4B}
& Original   & 0.245     & 0.100 & 0.067 & 0.100 & 0.089 & 0.692 & 0.425 & 0.277 \\
& GRPO       & 0.709 & 0.500 & 0.400 & 0.400 & 0.433 & 0.850 & 0.825 & 0.595 \\
& Difficulty-UCB & 0.803 & \textbf{0.567} & \textbf{0.533} & 0.400 & 0.499 & 0.846 & 0.875 & 0.644 \\
\rowcolor{gray!20}
& PGB (Ours)   & \textbf{0.811} & 0.533 & \textbf{0.533} & \textbf{0.533} & \textbf{0.533} & \textbf{0.852} & \textbf{0.975} & \textbf{0.683} \\

\midrule

\multirow{4}{*}{Qwen3-8B}
& Original   & 0.204     & 0.100 & 0.067 & 0.133 & 0.099 & 0.414 & 0.300 & 0.202 \\
& GRPO       & 0.682 & 0.467 & 0.400 & 0.467 & 0.445 & 0.870 & 0.850 & 0.611 \\
& Difficulty-UCB & 0.769 & 0.633 & \textbf{0.533} & 0.433 & 0.531 & 0.844 & \textbf{0.900} & 0.667 \\
\rowcolor{gray!20}
& PGB (Ours)   & \textbf{0.774} & \textbf{0.733} & 0.400 & \textbf{0.600} & \textbf{0.577} & \textbf{0.872} & \textbf{0.900} & \textbf{0.700} \\

\bottomrule
\end{tabular}
}
\caption{Accuracy comparison on ID and OOD benchmarks across different model scales. Our PGB consistently achieves the best or comparable results across settings compared with standard GRPO and difficulty-UCB curriculum baselines, demonstrating strong generalization and scalability.}
\label{tab:results}
\vspace{-0.4cm}
\end{table*}

%% file: Input/image_training_curve.tex
\begin{figure}[t]
    \centering
    \includegraphics[width=0.86\linewidth]{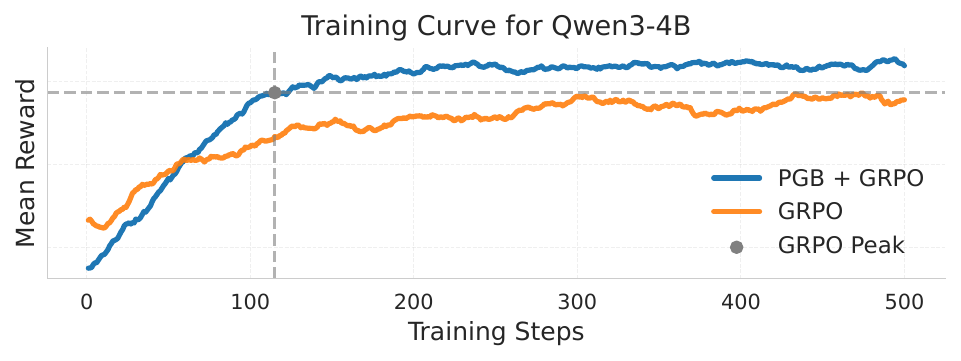}
    \vspace{0.1cm}
    \includegraphics[width=0.86\linewidth]{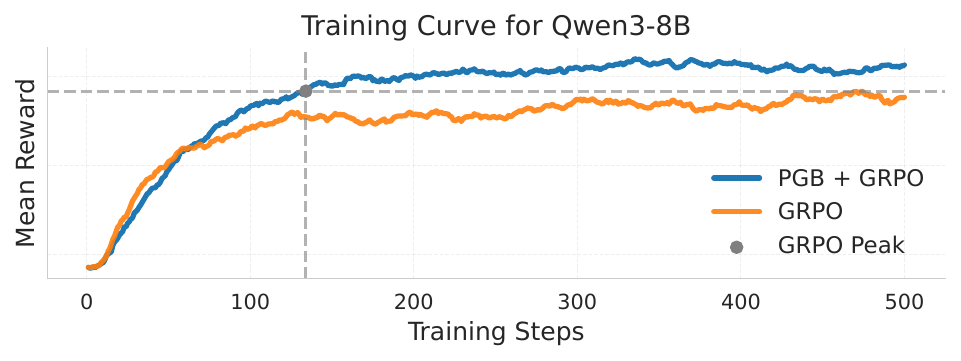}
    \vspace{-0.2cm}
    \caption{Training curves of mean reward for PGB + GRPO and GRPO on Qwen3-4B and 8B models. }
    
    \label{image:training_curve}
    \vspace{-0.4cm}

\end{figure}

%% file: Input/table_training_curve.tex
\begin{table}[t]
    \centering
    \begin{adjustbox}{max width=\columnwidth}
    \begin{tabular}{c|cc|cc}
\hline
Steps & 4B-PGB & 4B-GRPO & 8B-PGB & 8B-GRPO \\
\hline
100 & \textbf{0.741} & 0.628 & \textbf{0.699} & 0.621 \\
200 & \textbf{0.807} & 0.645 & \textbf{0.748} & 0.645 \\
300 & \textbf{0.791} & 0.695 & \textbf{0.762} & 0.669 \\
400 & \textbf{0.815} & 0.677 & \textbf{0.771} & 0.664 \\
500 & \textbf{0.811} & 0.709 & \textbf{0.771} & 0.682 \\
\hline
    \end{tabular}
    \end{adjustbox}
    \caption{Accuracy over steps on the DAPO dataset for PGB and standard GRPO on Qwen3-4B and 8B models.}
    \vspace{-0.2cm}
    \label{tab:training_cruve}
\end{table}

%% file: Input/table_unit_construct.tex
\begin{table}[t]
    \centering
    \setlength{\tabcolsep}{4pt}
    \begin{tabular}{ccccc}
\hline
Method & DAPO & AIME & MATH-500 & AMC \\
\hline
GRPO & 0.709 & 0.433 & 0.850 & 0.825 \\
Difficulty & 0.800 & 0.499 & 0.842 & 0.875 \\
Rubric & 0.807 & 0.433 & 0.846 & 0.900 \\
\rowcolor{gray!20}
Gradient & \textbf{0.811} & \textbf{0.533} & \textbf{0.852} & \textbf{0.975} \\
\hline
    \end{tabular}
    \caption{Comparison of different unit construction strategies across math benchmarks on Qwen3-4B.}
    \vspace{-0.4cm}
    \label{tab:construct}
\end{table}

%% file: Input/table_sampling.tex
\begin{table}[t]
    \centering
    \setlength{\tabcolsep}{4pt}
    \begin{tabular}{ccccc}
\hline
Method & DAPO & AIME & MATH-500 & AMC \\
\hline
Boltzmann & 0.570 & 0.242 & 0.702 & 0.800 \\
UCB & 0.572 & 0.222 & 0.706 & 0.775 \\
\rowcolor{gray!20}
Predictive & \textbf{0.608} & \textbf{0.322} & \textbf{0.710} & \textbf{0.800} \\
\hline
    \end{tabular}
    \caption{Comparison of predictive and reward-based curriculum scheduling strategies on reasoning benchmarks on Qwen3-1.7B.}
    \vspace{-0.35cm}
    \label{tab:sampling}
\end{table}

%% file: Input/image_unit_prob.tex
\begin{figure}[t]
    \centering
    \begin{subfigure}{\linewidth}
        \centering
        \includegraphics[width=0.9\linewidth]{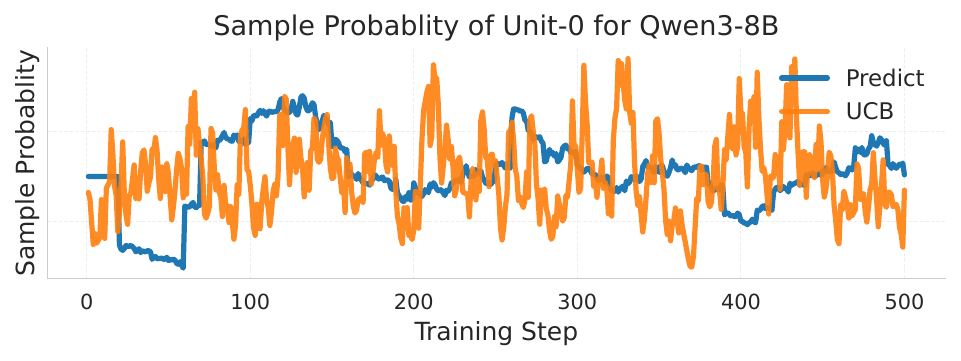}
        \vspace{-0.2cm}
        \caption{Sampling probability dynamics of a learning unit under predictive and UCB-based scheduling.}
        \label{fig:sampling_prob}
    \end{subfigure}

    \begin{subfigure}{\linewidth}
        \centering
        \includegraphics[width=0.9\linewidth]{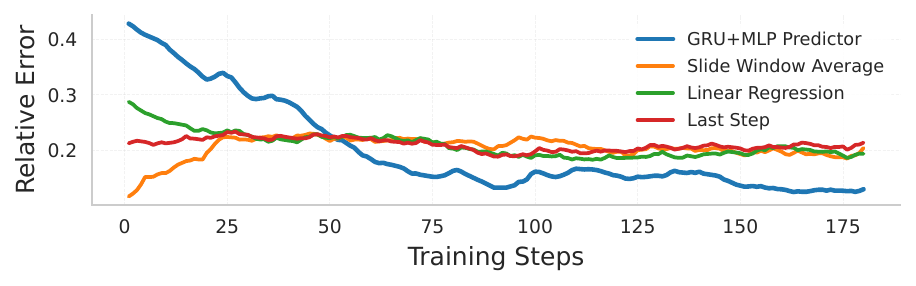}
        \vspace{-0.2cm}
        \caption{Comparison of different future utility prediction strategies.}
        \label{fig:compare_predict}
    \end{subfigure}
    \vspace{-0.35cm}
    \caption{Analysis of predictive curriculum scheduling.}
    
    \label{image:predict}
    \vspace{-0.35cm}

\end{figure}

%% file: Input/appendix.tex
\section{Predictive Bandit CRL Algorithm}
\input{Input/algorithm_small}

\section{Theoretical Explanation}
\subsection{First-Order Interaction between Sample Gradients}
For training samples $(x_i,y_i)$ and $(x_j,y_j)$, we 

Assume that $\ell_j(\theta)$ is twice differentiable and its Hessian is bounded by $\|\nabla^2_\theta \ell_j(\theta)\|_2 \leq H$. For the update $\theta'=\theta-\eta g_i$, let $\Delta\theta = \theta'-\theta=-\eta g_i$, Taylor's theorem gives
\begin{equation}
\begin{aligned}
    \ell_j(\theta') &= \ell_j(\theta)+\nabla_\theta \ell_j(\theta)^\top \Delta\theta+\frac{1}{2}
\Delta\theta^\top\nabla_\theta^2\ell_j(\tilde{\theta})\Delta\theta\\
    &=\ell_j(\theta)-\eta g_j^\top g_i+\frac{\eta^2}{2}
g_i^\top
\nabla_\theta^2\ell_j(\tilde{\theta})
g_i,
\end{aligned}
\end{equation}
Rearranging and taking absolute values, we obtain
\begin{equation}
\left|\ell_j(\theta')-\ell_j(\theta)+\eta g_j^\top g_i\right|
=\frac{\eta^2}{2}\left|g_i^\top\nabla_\theta^2\ell_j(\tilde{\theta})g_i\right|.
\end{equation}
Using the quadratic-form bound
\begin{equation}
|x^\top A x|\leq\|A\|_2\|x\|_2^2
\end{equation}
and the Hessian bound
\begin{equation}
\left\|\nabla_\theta^2\ell_j(\tilde{\theta})\right\|_2\leq H,
\end{equation}
we have
\begin{equation}
\left|\ell_j(\theta')-\ell_j(\theta)+\eta g_j^\top g_i\right|\leq\frac{\eta^2H}{2}\|g_i\|_2^2.
\end{equation}

\subsection{Properties of Gradient Random Projection}
To reduce the computational cost of clustering high-dimensional sample gradients, we apply a random projection before clustering. 
According to the Johnson-Lindenstrauss Lemma~\cite{johnson1984extensions}, given $0<\varepsilon<1$, a set $X\subset\mathbb{R}^D$ and a number $d>\Omega(\frac{\log D}{\varepsilon^2})$, there is a linear map $f:\mathbb{R}^D\to\mathbb{R}^d$ such that
\begin{equation}
    (1-\varepsilon)\|u-v\|^2\leqslant \left\|f(u)-f(v)\right\|^2\leqslant (1+\varepsilon)\|u-v\|^2
\end{equation}
for all $u,v\in X$. 
Specifically, given a normalized gradient representation $\bar{g}_i \in \mathbb{R}^{D}$, we project it into a lower-dimensional space using a random matrix $R \in \mathbb{R}^{d \times D}$:
\begin{equation}
z_i = R\bar{g}_i,
\end{equation}
where each entry of $R$ is independently sampled from a Gaussian distribution:

\begin{equation}
R_{mn}\sim\mathcal{N}(0,\frac{1}{d}).
\end{equation}

The projection approximately preserves pairwise distances and inner products according to the Johnson--Lindenstrauss lemma. Here, we show that the projected inner product is an unbiased estimator of the original gradient similarity.

For two normalized gradient representations $\bar{g}_i$ and $\bar{g}_j$, their projected similarity is:

\begin{equation}
\langle z_i,z_j\rangle = \langle R\bar{g}_i,R\bar{g}_j\rangle = \bar{g}_i^{\top}R^{\top}R\bar{g}_j .
\end{equation}
Since each element of $R$ follows
$\mathcal{N}(0,\frac{1}{d})$, we have:
\begin{equation}
\mathbb{E}[R^{\top}R]=I_D .
\end{equation}
Therefore,
\begin{equation}
    \begin{aligned}
\mathbb{E}[\langle z_i,z_j\rangle]
&=
\mathbb{E}
[\bar{g}_i^{\top}R^{\top}R\bar{g}_j]
\\
&=
\bar{g}_i^{\top}
\mathbb{E}[R^{\top}R]
\bar{g}_j
\\
&=
\bar{g}_i^{\top}\bar{g}_j .
\end{aligned}
\end{equation}

Thus, the expected inner product between projected gradients is equal to the original gradient similarity. With sufficiently large projection dimension $d$, the Johnson--Lindenstrauss lemma further guarantees that pairwise distances are preserved with high probability:

\begin{equation}
(1-\epsilon)\|\bar{g}_i-\bar{g}_j\|_2^2
\leq
\|z_i-z_j\|_2^2
\leq
(1+\epsilon)\|\bar{g}_i-\bar{g}_j\|_2^2 ,
\end{equation}
where $\epsilon$ controls the distortion introduced by the random projection.
Therefore, random projection preserves the relative geometry of sample gradients while substantially reducing the dimensionality, allowing efficient clustering of optimization-aware learning units.

\section{Implementation Details}
This section provides implementation details for reproducing our experiments, including the model and dataset configurations, learning unit construction, predictor architecture, curriculum scheduling, and RL hyperparameters. Unless otherwise specified, all experiments follow the same training protocol.

We implement PGB using the VERL reinforcement learning framework with GRPO. We train Qwen3-1.7B, Qwen3-4B-Thinking, and Qwen3-8B on DAPO-Math-17K. For evaluation, AIME 2024, 2025, and 2026 each contain 30 problems, while AMC and MATH-500 contain 40 and 500 problems, respectively. The training data comprise English mathematical reasoning problems from the English subset of DAPO-Math-17K. The maximum prompt and response lengths are 2,048 and 8,192 tokens, respectively. Each model is trained for 500 steps with a global batch size of 32. In the ablation studies, we vary only the component under investigation while keeping the remaining configurations fixed.

To construct learning units, we use Qwen3-30B to generate reference responses for the DAPO-Math-17K training prompts. We compute sample-level SFT gradients under the initial policy, normalize them, and project them to $d=1024$ dimensions before partitioning the training data into $K=8$ learning units using spherical K-means. We then treat the units as arms in a probabilistic bandit. Our curriculum scheduler uses a GRU followed by an MLP to predict each unit's next-step mean absolute advantage from historical learning states. Reward progress, sample counts, and gradient statistics serve as contextual features. Training begins with a 20-step warm-up stage using uniform sampling. Subsequently, the predictor is updated every 10 training steps, and its predictions guide sampling with a Boltzmann temperature of 0.4. At each predictor update, we perform 5 optimization iterations with a learning rate of $1\times10^{-3}$. The GRU and MLP hidden dimensions are 64 and 32, respectively, and the history sequence length is 10.

For GRPO optimization, the actor learning rate is $1\times10^{-6}$. We use dynamic batch sizing and gradient checkpointing. Rollouts are generated with vLLM using tensor parallelism of 4 and 8 responses per prompt. All experiments use FSDP-based distributed training on a single node with 8 NVIDIA H20 GPUs (141 GB per GPU). Due to the computational cost of RL post-training, we report one independent run per configuration.

\subsection{Gradient-Based Unit Construction}
Although the policy evolves during RL, we construct learning units only once using the initial policy. The resulting units serve as stable data partitions, while their stage-dependent usefulness is captured by the predictive scheduler. Re-clustering throughout training would introduce considerable overhead and change the arm identity, making temporal utility modeling difficult.

\input{Input/table_appendix_temperature}
\input{Input/image_appendix_gradient_ana}
\input{Input/table_appendix_task_grad}

\section{Experiments}
\subsection{Ablation Study on the Temperature Parameter $\tau$}
We conduct an ablation study to investigate the effect of the temperature parameter $\tau$, which controls the exploration-exploitation trade-off in the Boltzmann allocation strategy. The results in Table~\ref{tab:appendix_tau} indicate that PGB is robust to different temperature settings, while extreme values lead to slightly worse performance. A small $\tau$ makes the sampling distribution more concentrated, which may reduce exploration and limit the discovery of useful  learning units. In contrast, a large $\tau$ increases exploration but weakens the exploitation of high-utility units. Overall, $\tau=0.4$ achieves the best performance on most benchmarks, obtaining the highest scores on DAPO-Math, MATH-500, and AMC. Therefore, we use $\tau=0.4$ as the default setting.

\subsection{Gradient Analysis of Predefined Task Groups}
To investigate whether task-based grouping reflects the intrinsic optimization behaviors of training samples, we conduct an analysis on the DAPO-MATH dataset. Specifically, we categorize the dataset into eight task groups according to rubric annotations and uniformly sample 1,000 training examples. For each sample, we compute its gradient representation with respect to the model parameters and analyze the gradient characteristics within each task group.

As shown in Figure~\ref{image:appendix_gradient}, samples from the same task are widely distributed in the gradient space and substantially overlap with samples from other categories. This indicates that task-level semantic categories do not form coherent clusters according to model-induced optimization behaviors. In contrast, clustering samples based on gradient representations produces more compact and distinguishable groups, suggesting that gradient similarity better captures the intrinsic optimization structure of training data.

We further quantify the gradient characteristics of task-based groups in Table~\ref{tab:appendix_task_group_statistics}. The standard deviation of gradient norms varies across groups, with several groups (e.g., Group 6 and 7) exhibiting large internal gradient variance (0.907 and 1.100, respectively). These statistics indicate substantial within-group variation in gradient magnitudes, complementing the gradient-space visualization.

These observations demonstrate that predefined task fail to characterize optimization-consistent sample groups. Therefore, we construct learning units based on gradient similarity to obtain optimization-aware curriculum structures.

\input{Input/image_appendix_unit_prob}

\input{Input/table_science}
\subsection{Ablation Study on Utility Prediction Strategies}
To evaluate the effectiveness of the proposed utility prediction module, we compare four strategies for estimating future utility: 
\begin{enumerate}
    \item \textbf{GRU+MLP Predictor}, models the temporal evolution of utility by taking historical utility states as input and predicting the utility at the next training step.
    \item \textbf{Sliding Window Average}, estimates future utility using the average utility over recent steps. 
    \item \textbf{Linear Regression method}, fits a linear function over historical utility values to extrapolate future trends.
    \item \textbf{Last Step method}, directly uses the most recent utility value as the prediction.
\end{enumerate}

We train Qwen3-4B-Thinking on the DAPO-Math dataset using DAPO for 200 steps, including the warm-up stage,
and measure prediction accuracy using the average relative error of four future utility prediction strategies throughout the training process:
\begin{equation}
    \operatorname{RE}^{t}
    =
    \frac{1}{b}
    \sum_{k=1}^{b}
    \frac{
    \left|\hat{R}_k^{t}-R_k^{t}\right|
    }{
    \left|R_k^{t}\right|+\epsilon
    },
\end{equation}
where $\epsilon$ is a small constant for numerical stability.
The results are shown in Figure~\ref{image:appendix_err_predict}. The GRU+MLP Predictor initially shows higher errors due to the limited historical trajectories available at the early stage. As more training data are collected, its prediction error decreases continuously and becomes lower than all other approaches. We further summarize the prediction errors using the mean and median values. GRU+MLP Predictor achieves the lowest average and median errors among all strategies, with a mean error of 0.1698 and a median error of 0.1475. In comparison, the best-performing baseline obtains a mean error of 0.2000 and a median error of 0.1946. These results demonstrate that explicitly modeling temporal utility dynamics provides more accurate future utility estimation for curriculum scheduling.

Figure~\ref{image:appendix_unit_prob} further extends the sampling-dynamics visualization presented in the main text by reporting the allocation trajectories for learning units. Across units, predictive scheduling generally produces smoother probability trajectories than UCB-based scheduling, while preserving unit-specific changes in sampling priorities throughout training.

\subsection{Generalization to Science Reasoning Tasks}
While our main experiments focus on mathematical reasoning tasks, we further evaluate the generalization ability of PGB on science reasoning tasks. Specifically, we conduct experiments on SciKnowEval~\cite{feng2025sciknoweval}, covering Chemistry, Biology and Physics questions. We compare PGB with the original model and standard GRPO to investigate whether predictive curriculum scheduling remains effective beyond math reasoning scenarios. As shown in Table \ref{tab:science}, PGB consistently improves the performance across different science reasoning domains. For example, on the chemistry benchmark, PGB achieves an accuracy of $52.4\%$, outperforming standard GRPO by 1.9 percentage points. Similar improvements are observed on biology and physics tasks. These results provide evidence that PGB can extend beyond mathematical reasoning.

%% file: Input/algorithm_small.tex
\begin{algorithm}[ht]
\caption{Predictive Multi-armed Bandit CRL}
\label{alg:pmab}
\textbf{Input:} Dataset $\mathcal D$, learning units $\mathcal U=\{u_1,\dots,u_K\}$, policy $\theta$, warm-up steps $T_{\mathrm{warm}}$, predictor update interval $T_{pred}$\\
\textbf{Output:} Trained policy parameters $\theta$

\begin{algorithmic}[1]

\STATE Initialize predictor $f_\phi$

\vspace{0.3em}
\STATE \textbf{Stage 1: Warm-up}
\FOR{$t=1,\ldots,T_{\text{warm}}$}
    \STATE Uniformly sample training batches from $\{u_i\}$.

    \STATE Update policy $\theta$ using RL optimization.

    \STATE Collect learning states 
    $\mathcal{H}_k =\mathcal{H}_k \cup\{\mathbf{s}_k^t\}$.
\ENDFOR
\vspace{0.3em}

\STATE Train predictor $f_\phi$ using $\mathcal{H}_k^t$

\vspace{0.3em}
\STATE \textbf{Stage 2: Predictive Sampling}
\FOR{$t=T_{\text{warm}}+1,\ldots,T$}

    \STATE Predict future utility of all learning units:
    
    $
    \qquad\{\hat R_k^{t+1}\}_{k=1}^{K}
    =
    f_\phi(\mathcal H_1^t,\dots,\mathcal H_K^t)
    $

    \STATE Compute sampling distribution $\pi^{t+1}$ 

    \STATE Sample training batch according to $\pi^{t+1}$.

    \STATE Update policy $\theta$ using RL optimization.

    \STATE Update $f_\phi$ every $T_{pred}$ steps.

\ENDFOR

\STATE \textbf{return} $\theta$

\end{algorithmic}

\end{algorithm}
%\vspace{-0.5cm}

%% file: Input/table_appendix_temperature.tex
\begin{table}[t]
    \centering
    \setlength{\tabcolsep}{4pt}
    \begin{tabular}{ccccc}
\hline
$\tau$ & DAPO & AIME & MATH-500 & AMC \\
\hline
0.2 & 0.739 & 0.489 & 0.854 & 0.825 \\
0.3 & 0.740 & 0.411 & 0.860 & 0.875 \\
0.4 & \textbf{0.761} & 0.489 & \textbf{0.862} & \textbf{0.95} \\
0.5 & 0.745 & 0.489 & 0.860 & \textbf{0.95}\\
0.6 & 0.734 & \textbf{0.500} & 0.860 & 0.9 \\
\hline
    \end{tabular}
    \caption{Comparison of different temperature parameter $\tau$ across math reasoning benchmarks on Qwen3-4B.}
    \vspace{-0.3cm}
    \label{tab:appendix_tau}
\end{table}

%% file: Input/image_appendix_gradient_ana.tex
\begin{figure}[t]
    \centering
    \includegraphics[width=0.9\linewidth]{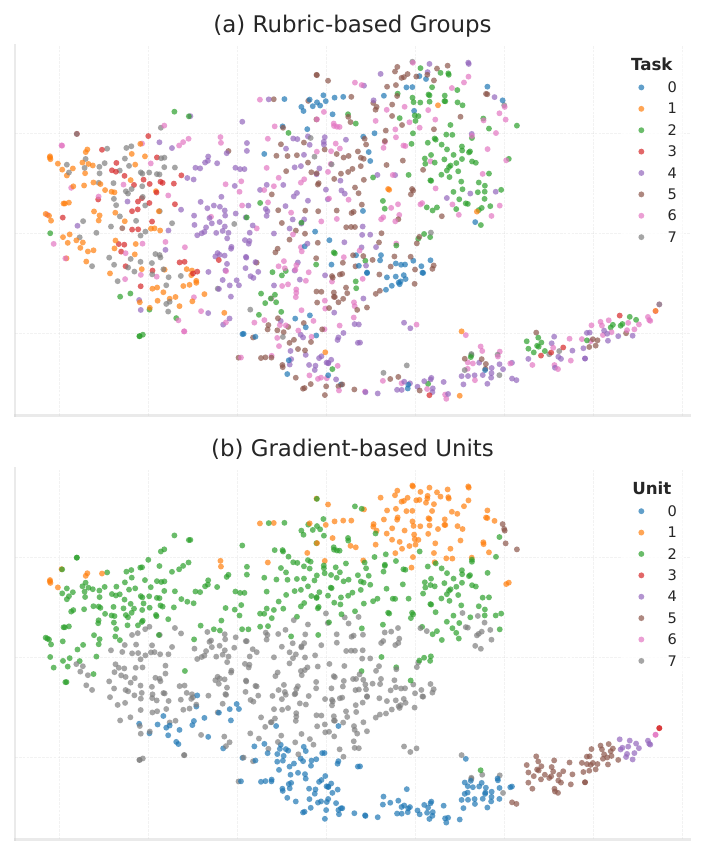}
    \caption{Visualization of task-based groups and gradient-based learning units in gradient space. Samples from the same rubric category are widely distributed and overlap with other categories, while gradient-based clustering discovers more coherent optimization structures.}
    
    \label{image:appendix_gradient}
    \vspace{-0.2cm}

\end{figure}

%% file: Input/table_appendix_task_grad.tex
\begin{table}[t]
\centering

\begin{tabular}{c|ccc}
\toprule
Group & Grad Norm Std & Loss Mean & Loss Std \\
\midrule
0 & 0.567 & 0.731 & 0.118 \\
1 & 0.657 & 0.838 & 0.109 \\
2 & 0.520 & 0.861 & 0.125 \\
3 & 0.496 & 0.772 & 0.116 \\
4 & 0.576 & 0.707 & 0.148 \\
5 & 0.567 & 0.729 & 0.129 \\
6 & 0.907 & 0.778 & 0.155 \\
7 & 1.100 & 0.795 & 0.100 \\
\bottomrule

\end{tabular}
\caption{Optimization statistics of task-based groups.}
\label{tab:appendix_task_group_statistics}
\end{table}

%% file: Input/image_appendix_unit_prob.tex
\begin{figure}[t]
    \centering
    \includegraphics[width=0.95\linewidth]{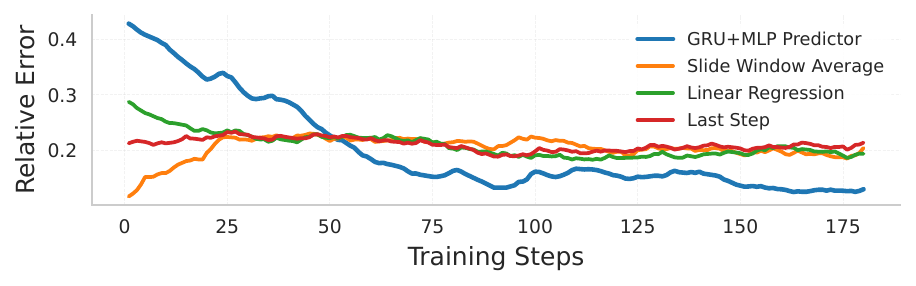}
    \vspace{-0.3cm}
    \caption{Comparison of future utility prediction strategies.}
    
    \label{image:appendix_err_predict}

\end{figure}

\begin{figure}[t]
    \centering
    \includegraphics[width=0.9\linewidth]{Figures/prob_0.pdf}
    \includegraphics[width=0.9\linewidth]{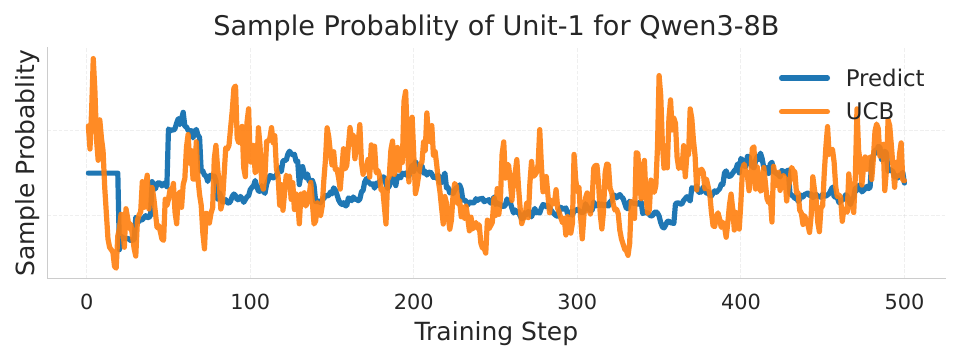}
    \includegraphics[width=0.9\linewidth]{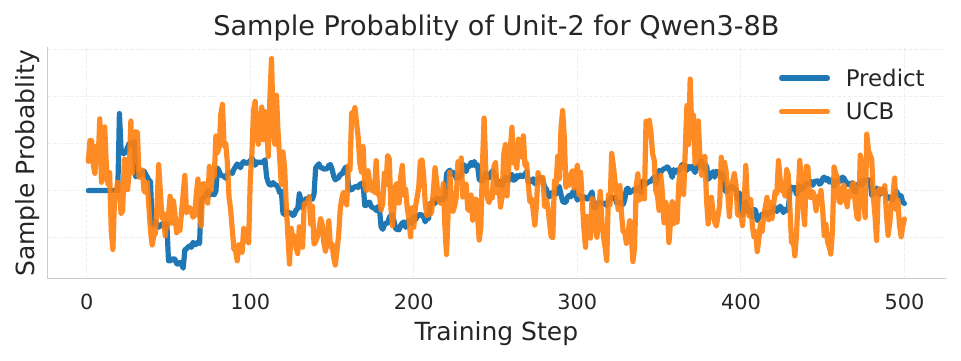}
    \includegraphics[width=0.9\linewidth]{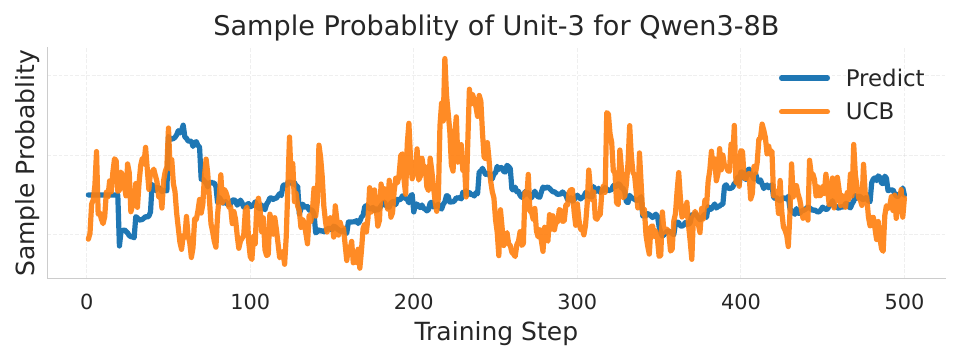}
    \includegraphics[width=0.9\linewidth]{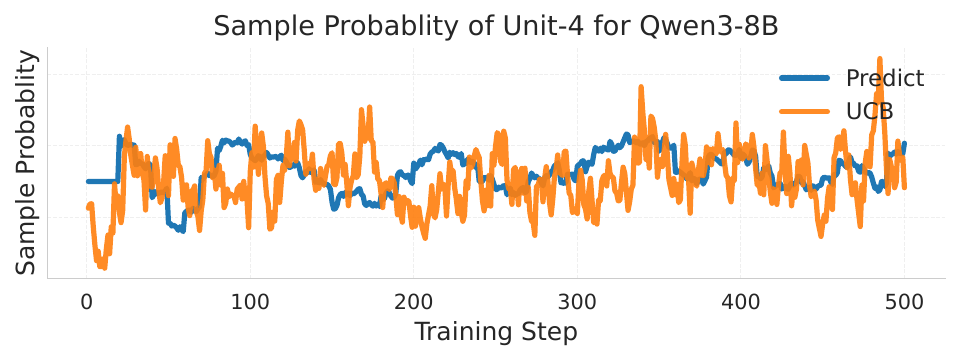}
    \vspace{-0.3cm}
    \caption{Sampling probability dynamics of learning unit 0-4 under predictive and UCB-based curriculum scheduling.}
    
    \label{image:appendix_unit_prob}
    \vspace{-0.3cm}

\end{figure}

%% file: Input/table_science.tex
\begin{table}[t]
    \centering
    \setlength{\tabcolsep}{6pt}
    \begin{tabular}{cccc}
\hline
Method & Chemistry & Biology & Physics \\
\hline
Original & 0.443 & 0.240 & 0.500 \\
GRPO & 0.505 & 0.260 & 0.525 \\
PGB & \textbf{0.524} & \textbf{0.300} & \textbf{0.538} \\
\hline
    \end{tabular}
    \caption{Performance comparison on science reasoning tasks from SciKnowEval~\cite{feng2025sciknoweval} with Qwen3-1.7B.}
    \vspace{-0.5cm}
    \label{tab:science}
\end{table}